\documentclass{article}

\usepackage[preprint]{neurips_2023}

\usepackage[utf8]{inputenc} 
\usepackage[T1]{fontenc}    
\usepackage{url}            
\usepackage{booktabs}       
\usepackage{amsfonts}       
\usepackage{nicefrac}       
\usepackage{microtype}      
\usepackage{xcolor}         

\usepackage{times}
\usepackage{epsfig}
\usepackage{graphicx}
\usepackage{amsmath}
\usepackage{amssymb}

\usepackage{xcolor}
\usepackage{algorithm}
\usepackage{listings}
\usepackage[british,american]{babel}
\usepackage{enumitem}
\usepackage{multirow}

\usepackage{textpos}  

\usepackage{amsmath,amsfonts,bm}
\usepackage{xcolor}
\usepackage{fontawesome}

\def\eqref#1{(\ref{#1})}

\def\1{\bm{1}}

\def\vx{{\bm{x}}}
\def\vy{{\bm{y}}}

\def\mA{{\bm{A}}}

\def\mG{{\bm{G}}}

\def\mJ{{\bm{J}}}

\def\mS{{\bm{S}}}

\def\mX{{\bm{X}}}
\def\mY{{\bm{Y}}}

\DeclareMathAlphabet{\mathsfit}{\encodingdefault}{\sfdefault}{m}{sl}
\SetMathAlphabet{\mathsfit}{bold}{\encodingdefault}{\sfdefault}{bx}{n}

\def\sN{{\mathbb{N}}}

\newcommand{\R}{\mathbb{R}}

\newcommand{\abs}[1]{\left\lvert #1 \right\rvert}

\newcommand{\norm}[1]{\left\lVert#1\right\rVert}
\newcommand{\T}{\top}

\newcommand{\Lip}{{\rm Lip}}

\usepackage{stfloats}
\usepackage{float}
\usepackage{siunitx}
\usepackage{caption}
\usepackage{subcaption}
\usepackage{amsthm}
\newtheorem{definition}{Definition}
\usepackage[most]{tcolorbox}
\usepackage{listings}
\usepackage{wrapfig,lipsum}
\usepackage{pifont}
\usepackage{xspace}
\usepackage{soul}
\usepackage{etoc}
\usepackage{colortbl}
\usepackage{bbm}
\usepackage{algorithm}
\usepackage{xcolor}
\usepackage{url}
\usepackage{csquotes}
\usepackage{graphicx}
\usepackage{amsmath}
\usepackage{mathtools}
\usepackage{amssymb}
\usepackage{amsfonts}
\usepackage{bm}
\usepackage{booktabs}
\usepackage{bbm}
\usepackage{multirow}
\usepackage{enumitem}
\usepackage{wrapfig,lipsum}
\usepackage{algorithm}
\usepackage{xcolor}
\usepackage{rotating}
\usepackage{makecell}
\usepackage{amsmath,amsfonts,bm}
\usepackage{natbib}
\setcitestyle{numbers,square}
\usepackage{listings}
\usepackage{colortbl}
\definecolor{mydarkblue}{rgb}{0,0.08,0.45}
\definecolor{myblue}{HTML}{3b75c3}
\definecolor{myred}{HTML}{E33222}
\definecolor{mygreen}{HTML}{438773}
\definecolor{mymaroon}{RGB}{142,27,19}
\definecolor{maroon}{HTML}{800000}
\definecolor{mycite}{cmyk}{0.55,1,0,0.15}
\definecolor{codeblue}{rgb}{0.25,0.5,0.5}
\definecolor{codekw}{rgb}{0.85, 0.18, 0.50}
\definecolor{codegreen}{rgb}{0,0.6,0}
\definecolor{codegray}{rgb}{0.5,0.5,0.5}
\definecolor{codepurple}{rgb}{0.58,0,0.82}
\definecolor{backcolour}{rgb}{0.95,0.95,0.92}
\definecolor{mygray}{gray}{0.925}
\definecolor{beaublue}{HTML}{fefae0}
\definecolor{blackish}{rgb}{0.2, 0.2, 0.2}

\usepackage{url}
\usepackage{graphicx}
\usepackage{csquotes}
\usepackage{graphicx}
\usepackage{amsmath}
\usepackage{mathtools}
\usepackage{amssymb}
\usepackage{amsfonts}
\usepackage{bm}
\usepackage{booktabs}
\usepackage{bbm}
\usepackage{multirow}
\usepackage{enumitem}
\usepackage{wrapfig,lipsum}
\usepackage{algorithm}
\usepackage{xcolor}
\usepackage{rotating}
\usepackage{makecell}
\usepackage{listings}
\usepackage{colortbl}
\usepackage{pifont}
\usepackage{xspace}
\usepackage{soul}
\usepackage{etoc}

\newtheorem{lemma}{Lemma}
\newtheorem{theorem}{Theorem}

\newcolumntype{a}{>{\columncolor{mygray}}r}

\definecolor{mygray}{gray}{0.6}

\newcommand{\app}{\raise.17ex\hbox{$\scriptstyle\sim$}}

\usepackage{tabulary}
\newcolumntype{x}[1]{>{\centering\arraybackslash}p{#1pt}}
\usepackage{etoolbox}
\makeatletter
\AfterEndEnvironment{algorithm}{\let\@algcomment\relax}
\AtEndEnvironment{algorithm}{\kern2pt\hrule\relax\vskip3pt\@algcomment}
\let\@algcomment\relax
\newcommand\algcomment[1]{\def\@algcomment{\footnotesize#1}}
\renewcommand\fs@ruled{\def\@fs@cfont{\bfseries}\let\@fs@capt\floatc@ruled
  \def\@fs@pre{\hrule height.8pt depth0pt \kern2pt}%
  \def\@fs@post{}%
  \def\@fs@mid{\kern2pt\hrule\kern2pt}%
  \let\@fs@iftopcapt\iftrue}
\makeatother

\definecolor{citecolor}{HTML}{0071bc}
\definecolor{mathcolor}{HTML}{0071bc}
\usepackage[pagebackref=true,breaklinks=true,colorlinks,citecolor=citecolor,bookmarks=false]{hyperref}
\usepackage[capitalize,noabbrev,nameinlink]{cleveref}

\title{Promoting Fairness in GNNs: A Characterization of Stability}

\author{Yaning Jia$^{1}$,$^{*}$ Chunhui Zhang$^2$\thanks{Both authors ($^1$jiayaning@hust.edu.cn $^2$chunhui.zhang.gr@dartmouth.edu) contributed equally in alphabetical order. }\\
$^1$Huazhong University of Science and Technology, $^2$Dartmouth College \\
}

\begin{document}

\maketitle

\begin{abstract}
The Lipschitz bound, a technique from robust statistics, can limit the maximum changes in the output concerning the input, taking into account associated irrelevant biased factors. It is an efficient and provable method for examining the output stability of machine learning models without incurring additional computation costs. 
Recently, Graph Neural Networks (GNNs), which operate on non-Euclidean data, have gained significant attention. 
However, no previous research has investigated the GNN Lipschitz bounds to shed light on stabilizing model outputs, especially when working on non-Euclidean data with inherent biases. 
Given the inherent biases in common graph data used for GNN training, it poses a serious challenge to constraining the GNN output perturbations induced by input biases, thereby safeguarding fairness during training.
Recently, despite the Lipschitz constant's use in controlling the stability of Euclidean neural networks, the calculation of the precise Lipschitz constant remains elusive for non-Euclidean neural networks like GNNs, especially within fairness contexts.
To narrow this gap, we begin with the general GNNs operating on an attributed graph, and formulate a Lipschitz bound to limit the changes in the output regarding biases associated with the input. 
Additionally, we theoretically analyze how the Lipschitz constant of a GNN model could constrain the output perturbations induced by biases learned from data for fairness training. 
We experimentally validate the Lipschitz bound's effectiveness in limiting biases of the model output. 
Finally, from a training dynamics perspective, we demonstrate why the theoretical Lipschitz bound can effectively guide the GNN training to better trade-off between accuracy and fairness. {{This \textbf{version} contains supplementary details.}}
\end{abstract}

\section{Introduction}
Graphs, as a form of non-Euclidean data, are pervasive in numerous real-world applications, including recommender systems~\citep{shalaby2017help,huang2021graph,li2021on}, drug discovery~\citep{takigawa2013graph,li2017learning}, and knowledge engineering~\citep{rizun2019knowledge,wang2018acekg}, among others. Accompanying the surge in graph data volume, the need to analyze it has spurred the development of learning on non-Euclidean data that combine Graph Neural Networks (GNNs) with deep learning~\citep{Gori2005,Scarselli2005,li2015gated,hamilton2017inductive,xu2018powerful}. Graph Convolutional Networks (GCNs)~\citep{kipf2016semi,zhang2018link,fan2019graph}, which are the most referenced GNN architecture, utilize convolutional layers and message-passing mechanisms to enable learning on graphs.

{In parallel with the successful applications of GNNs across diverse real-world scenarios, there has been an increasing societal concern, prompting the development of prosocial/safe graph learning algorithms~\citep{dong2021individual, pfr, kang2020inform, mujkanovic2022are}. However, numerous existing works fall short in terms of the interpretability of the \textit{interactions between the GNN model and in the graph data training}. This ignorance impedes the control of the model parameters, as it usually fails to account for irrelevant factors hidden in the graph data, ultimately undermining the reliability of the model after training. In light of this, we aim to deepen our comprehension of GNNs and their input-output patterns in relation to model parameters. Consequently, we pose the following fundamental question:}
\begin{center}
\textit{Without extra computations, how can we constrain the unwanted changes in GNN output, \\ particularly when the graph training data has hidden learnable biases?}
\end{center}

This question seeks to address situations where the training graph data includes irrelevant statistics that lead to inappropriate and unfair correlations. By providing a tool that constrains perturbations in predictions of the GNN model during training, we can ensure that these predictions are not overly sensitive to correlations with unfair elements in the learning process. Such a solution could have a range of potential extensions, including improving GNN generalization, debiasing unfair factors induced by data biases in the GNN, and defending against perturbations on graph data by regularizing the model's output changes using its Lipschitz bounds.
Yet, despite the success of current methodologies for training GNNs, they typically offer limited interpretability regarding the relationship between the graph input and the predicted output. In particular, most existing works \textit{lack} a clear understanding of how biases, learned layer-by-layer from the input, ultimately affect output stability with fairness considerations~\citep{dong2021individual, kang2020inform, liao2021information, li2021on}. 

In the regime of robust statistics, Lipschitz bounds have been introduced to study the maximal possible changes in output corresponding to irrelevant (biased) factors of the input. Similarly, from the perspective of the rank-based individual fairness~\citep{dong2021individual}, the output is considered to satisfy individual fairness, 
{if the relative order of instances, as captured by two ranking lists (based on the similarity between instance $i$ and other instances in descending order), while remaining consistent between the input and output predictions, unaffected by hidden irrelevant biases in the input data.}
By introducing the Lipschitz bound to GNNs training, we can provide an interpretable solution to ensure fairness.

The goal of this study is to establish the Lipschitz bound for GNNs, {and then facilitate its calculation through a simplified path that regularizes model training to enhance output stability. As the initial exploration in this direction, we expand its applications to learning for rank-based individual fairness. This is motivated by the property that the constraint effects of Lipschitz bounds, which promote input-output consistency (in terms of ranking), align with the definition of rank-based individual fairness.}
Our contributions can be summarized as follows:
\begin{itemize}[leftmargin=*]
    \item We derived the Lipschitz bound for general GNNs. 
    Based on that, we characterize the GNN's prediction regarding perturbations induced by input biases, especially on individual fairness from a ranking perspective, then constrain the 
    possible unfair changes for stabilizing GNN models.
    \item We further facilitate the calculation of Lipschitz bounds by deriving the Jacobian matrix of the model for practical training. This allows for straightforward approximation and implementation.
    \item We show that our theoretical analysis of the Lipschitz for GNNs can guide experimental implementations for GNN fairness training. Our solution is plug-and-play, beneficial to different existing methods, and also effective on diverse datasets and various graph tasks as validated empirically.
\end{itemize}

\section{Preliminaries}
\label{sec:preliminaries}

\paragraph{Lipschitz Constant}
Let us start by reviewing some definitions related to Lipschitz functions. A function $f:\mathbb{R}^{n} \rightarrow \mathbb{R}^{m}$ is established to be Lipschitz continuous on an input set $\mathcal{X}\subseteq  \mathbb{R}^{n}$ if there exists a constant $K \geq 0$ such that for all $\vx,\vy \in \mathcal{X}$, $f$ satisfies the following inequality:
\begin{equation}
\label{eq-1}
    \norm{f(\vx) - f(\vy)} \le K\norm{\vx-\vy}, \forall \vx,\vy \in \mathcal{X}.
\end{equation}
The smallest possible $K$ in Equation~(\ref{eq-1}) is the Lipschitz constant of $f$, denoted as ${\rm Lip}(f)$:
\begin{equation}
\label{eq-2}
    {\rm Lip}(f) = \sup\limits_{\vx,\vy\in \mathcal{X}, \vx\ne \vy}\frac{\norm{f(\vx)-f(\vy)}}{\norm{\vx-\vy}},
\end{equation}
and we say that $f$ is a $K$-Lipschitz function.
\textit{The Lipschitz constant of a function is essentially the largest possible change of the output corresponding to a perturbation of the input of the unit norm.} This makes the Lipschitz constant of a neural network an important measure of its stability with respect to the input features. However, finding the exact constant can be challenging, so obtaining an upper bound is often the approach taken. Such an upper bound is called a Lipschitz bound.

\paragraph{Graph Neural Networks} We assume a given graph $G=G(V,E)$, where $V$ denotes the set of nodes and $E$ denotes the set of edges. We will use $\mX=\left\{\vx_{1},\vx_{2},\cdots,\vx_{N}\right\} \subset \mathbb{R}^{F}$ to denote the $N$ node features in $\R^F$, {as the input of any layer of a GNN}. By abuse of notation, when there is no confusion, we also follow GNN literature and consider $\mX$ as the $\R^{N \times F}$ matrix whose $i$-th row is given by $\vx_i^\T$, $i=1,\cdots,N$, though it unnecessarily imposes an ordering of the graph nodes.

GNNs are functions that operate on the adjacency matrix $\mA \in \mathbb{R}^{N \times N}$ of a graph $G$. Specifically, an $L$-layer GNN can be defined as a function $f:\mathbb{R}^{N \times F^{\rm in}}\to \mathbb{R}^{N \times F^{\rm out}}$ that depends on $\mA$. Formally, we adopt the following definition:
\begin{definition}
\label{def-2}
An $L$-layer GNN is a function $f$ that can be expressed as a composition of $L$ message-passing layers $h^{l}$ and $L-1$ activation functions $\rho^{l}$, as follows:
\begin{equation}\label{eq-23}
f = h^{L} \circ \rho^{L-1} \circ \cdots \circ \rho^{1} \circ h^{1} ,
\end{equation}
where $h^{l}: \mathbb{R}^{F^{l-1}} \to \mathbb{R}^{F^{l}}$ is the $l$-th message-passing layer, $\rho^{l}: \mathbb{R}^{F^{l}}\to \mathbb{R}^{F^{l}}$ is the non-linear activation function in the $l$-th layer, and 
$F^{l-1}$ and $F^{l}$ 
denote the input and output feature dimensions for the $l$-th message-passing layer $h^{l}$
, respectively. In addition, we set $l = 1, \cdots, L$.
\end{definition}

\paragraph{Rank-based Individual Fairness of GNNs}
\label{sec:rank-based-fairness}
Rank-based Individual Fairness on GNNs~\citep{dong2021individual} focuses on the relative ordering of instances rather than their absolute predictions. It ensures that similar instances, as measured by the similarity measure $S(\cdot,\cdot)$, receive consistent rankings or predictions. The criterion requires that if instance $i$ is more similar to instance $j$ than to instance $k$, then the predicted ranking of $j$ should be higher than that of $k$, \textit{consistently}. This can be expressed as:
\begin{equation}
\text{if } S(\vx_i, \vx_j) > S(\vx_i, \vx_k), \text{ then } \mY_{ij} > \mY_{ik},
\end{equation}
where $\mY_{ij}$ and $\mY_{ik}$ denote the predicted rankings or predictions for instances $\vx_j$ and $\vx_k$, respectively, based on the input instance $\vx_i$. The criterion ensures that the predicted rankings or predictions align with the relative similarities between instances, promoting fairness and preventing discriminatory predictions based on irrelevant factors.

\section{Estimating Upper Bounds of the Lipschitz Constants on GNNs for Fairness}
In this section, we estimate the upper bounds of the Lipschitz constants for GNNs, which is crucial for analyzing output perturbations induced by input biases: Initially, in Section~\ref{sec:stability}, we establish Lipschitz bounds for GNNs and provide closed-form formulas for these bounds. Next, in Section~\ref{sec:jaco-matrix}, we derive the model's Jacobian matrix to facilitate Lipschitz bounds calculation in practice. Lastly, in Section~\ref{sec:alg}, we use these bounds to explore individual fairness, demonstrating how model stability can ensure output consistency, aligning with the rank-based individual fairness definition.

\subsection{Stability of the Model Output}
\label{sec:stability}
A GNN is a function that transforms graph features from the input space to the output space. Formally, a GNN can be represented as ${f}: \mX \in \R^{N \times F^{\text{in}}} \rightarrow \mY \in \R^{N \times F^{\text{out}}}$, where $\mX$ is the input feature matrix, $\mY$ is the output feature matrix, and $N$ is the number of nodes in the graph.
To analyze the stability of the model's output, we examine the Lipschitz bound of the Jacobian matrix of the GNN model. In this regard, we introduce the following lemma that establishes an inequality relation:
\begin{tcolorbox}[width=1.0\linewidth, colback=citecolor!2.5, colframe=black, arc=0pt, boxsep=0mm, arc=2mm, left=2mm, right=2mm, top=2mm, bottom=2mm]
\begin{lemma}
\label{lemma:inequality}
For any vectors \(x\) and \(y\) in a Euclidean space, let \(g(x)\) and \(g(y)\) be vector-valued functions of \(x\) and \(y\) respectively, and let \(g_i\) be the \(i\)-th component function of \(g\). Then, we have the following inequality:
\begin{equation}
\label{lemma-eq-2}
    \frac{\left\|g(x)-g(y)\right \|}{\|x-y\|} \leqslant \left\| \left[\frac{g_i(x)-g_i(y)}{\|x-y\|}\right]_{i=1}^n \right\|,
\end{equation}PFR
where $\left\|\cdot\right \|$ represents the norm of a vector.
\end{lemma}
\end{tcolorbox}
Lemma \ref{lemma:inequality} provides an inequality that relates the norm of the difference between two vector-valued functions, \(g(x)\) and \(g(y)\), to the norm of a vector composed of the component-wise differences of the functions evaluated at \(x\) and \(y\). Based on Lemma \ref{lemma:inequality}, we can now present the following theorem:
\begin{tcolorbox}[width=1.0\linewidth, colback=citecolor!2.5, colframe=black, arc=0pt, boxsep=0mm, arc=2mm, left=2mm, right=2mm, top=5mm, bottom=2mm]
\vspace{-0.3cm}
\begin{theorem}
\label{thm-1}
Let $\mY$ be the output of an $L$-layer GNN (represented in $f(\cdot)$) with $\mX$ as the input. Assuming the activation function (represented in $\rho(\cdot)$) is ReLU with a Lipschitz constant of ${\rm Lip}({\rho}) = 1$, then the global Lipschitz constant of the GNN, denoted as ${\rm Lip}({f})$, satisfies the following inequality:
\begin{equation}
\label{eq:thm-1}
{\rm Lip}({f}) \leqslant \max_j \prod_{l=1}^{L} \left\| {F^{l}}^{\prime} \right\| \left\| \left[\mathcal{J}({h}^{l}) \right]_{j}\right\|_{\infty},
\end{equation}
where ${F^{l}}^{\prime}$ represents the output dimension of the $l$-th message-passing layer, $j$ is the index of the node (e.g., $j$-th), and the vector $\left[\mathcal{J}({h}^{l}) \right] = \left[\norm{\mJ_{1}({h}^{l})}, \norm{\mJ_{2}({h}^{l})}, \cdots, \norm{\mJ_{{F^{l}}^{\prime}}({h}^{l})}\right]$. Notably, $\mJ_i(h^{l})$ denotes the $i$-th row of the Jacobian matrix of the $l$-th layer's input and output, and $\left[\mathcal{J}({h}^{l}) \right]_{j}$ is the vector corresponding to the $j$-th node in the $l$-th layer ${h}^{l}(\cdot)$.
\end{theorem}
\end{tcolorbox}

\textit{Proof Sketch:}\footnote{Please refer to the Appendix~\ref{app:proof} for the complete proof due to space constraints.} We initiate with investigating the Lipschitz property of GNNs, considering $1$-Lipschitz activation functions (e.g., the widely-used ReLU~\citep{nair2010rectified}. The hidden states of node feature $x_1$ and $x_2$ are represented as $z_1$ and $z_2$ respectively. The Lipschitz constant between these hidden states is calculated as $\frac{\left\|z_{1} - z_{2}\right\|}{\left\|x_1 - x_2\right\|} = \frac{\|\left({h}(x_1) - {h}(x_2)\right)\|}{\left\|x_1 - x_2\right\|}$.
Using the triangle inequality, this equation is further bounded by $\frac{\left\|z_{1} - z_{2}\right\|}{\left\|x_1 - x_2\right\|} \leqslant \left\|\left[\frac{{h}(x_1)_{i} - {h}(x_2)_{i}}{\left\|x_1 - x_2\right\|}\right]_{i=1}^{F^{\prime}}\right\|.
$
The Lipschitz constant of individual elements is also analyzed, again using the triangle inequality. The proof then proceeds to analyze the Lipschitz constant of the individual elements using the operation of the $l$-th layer in ${f}(\cdot)$.
Finally, the Lipschitz constant for the GNN is established as ${\rm Lip}({f}) = \max_{j}\prod_{l=1}^{L} \left\| {F^{l}}^{\prime} \right\| \left\| \left[\mathcal{J}({h}^{l}) \right]_j\right\|_{\infty}.$
It establishes that the difference in the output $\mY$ is controlled by the Lipschitz constant of the GNN, ${\rm Lip}({f})$, and the difference in the input $\mX$. The above analysis is crucial for assessing the model output stability with respect to irrelevant biases learned layer-by-layer from the input, during the forward.

The aforementioned Theorem \ref{thm-1} bounds the global Lipschitz constant of the GNN based on the layer outputs and the corresponding Jacobian matrices. It establishes that the Lipschitz constant of a GNN regulates the magnitude of changes in the output induced by input biases, consequently guaranteeing the model output stability and fairness against irrelevant factors. 

\subsection{Simplifying Lipschitz Bounds Calculation of the GNN via Jacobian Matrix}
\label{sec:jaco-matrix}
The approach presented in Theorem~\ref{thm-1} for estimating the Lipschitz constants ${\rm Lip}({f})$ across different layers in the GNN ${f}(\cdot)$ requires unique explicit expressions for each component. This makes the process somewhat challenging. Nevertheless, this difficulty can be mitigated by computing the corresponding Jacobian matrices, as shown in Equation~(\ref{eq:thm-1}). By leveraging the values of each component in the Jacobian matrix, we can approximate the Lipschitz constants in a straightforward manner. Additionally, the expression $\prod_{l=1}^{L} \left\| {F^{l}}^{\prime} \right\| \left\| \left[\mathcal{J}({h}^{l}) \right]_{j}\right\|_{\infty}$ from Equation~(\ref{eq:thm-1}) provides valuable insights into the factors that influence the Lipschitz bounds, such as the dimensions of the output layer and the depth of the GNN layers.

However, considering the potential for multiple hierarchical layers in the network, their cumulative effect could lead to a significant deviation from the original bounds. Therefore, it is beneficial to consider the entire network as a single model and directly derive the Lipschitz bound from the input and output. In this subsection, we provide a \textit{simplified} method to derive the Lipschitz bound in Equation~(\ref{eq:thm-1}) for facilitating fairness training.

To achieve this, we introduce the Jacobian matrix in detail. Let $\left[\mJ_i\right]$ denote the Jacobian matrix of the $i$-th node, which can be calculated as:
\begin{equation}
\label{eq-4-1}
    \left[\mJ_{i}\right]_{F^{\text {out}} \times F^{\text{in}}}=
    \begin{bmatrix}
    \frac{\partial \mY_{i 1}}{\partial \mX_{i 1}}&  \frac{\partial \mY_{i 1}}{\partial \mX_{i 2}}  &\cdots&\frac{\partial \mY_{i 1}}{\partial \mX_{i F^{\text{in}}}} \\
    \frac{\partial \mY_{i 2}}{\partial \mX_{i 1}}&  \frac{\partial \mY_{i 2}}{\partial \mX_{i 2}}  &\cdots&\frac{\partial \mY_{i 2}}{\partial \mX_{i F^{\text{in}}}} \\
    \vdots     &  \vdots     &\vdots  &\vdots \\
    \frac{\partial \mY_{i F^{\text{out}}}}{\partial \mX_{i 1}}&  \frac{\partial \mY_{i F^{\text{out}}}}{\partial \mX_{i 2}}  &\cdots&\frac{\partial \mY_{i F^{\text{out}}}}{\partial \mX_{i F^{\text{in}}}}
    \end{bmatrix}_{F^{\text{out}} \times F^{\text{in}}}.
\end{equation}

We can define $\left[\mJ_{i}\right]_{F^{\text{out}} \times F^{\text{in}}} = \left[\mJ_{i 1}^{\top}, \mJ_{i 2}^{\top}, \cdots, \mJ_{i F^{\text{out}}}^{\top}\right]^{\top}$, where $\mJ_{i j} = \left[ \frac{\partial \mY_{i j}}{\partial \mX_{i 1}}, \frac{\partial \mY_{i j}}{\partial \mX_{i 2}}, \cdots, \frac{\partial \mY_{i j}}{\partial \mX_{i F^{\text{in}}}}  \right]^{\top}$. Then we let $\mathcal{J}_{i}=\left[ \norm{\mJ_{i 1}}, \norm{\mJ_{i 2}}, \cdots, \norm{\mJ_{i F^{\text{out}}}} \right]^{\top}$, $\mathcal{J} = \left[\mathcal{J}_{1}^{\top}, \mathcal{J}_{2}^{\top}, \cdots, \mathcal{J}_{N}^{\top} \right]^{\top}$. To analyze the Lipschitz bounds of the Jacobian matrix of all output features for $N$ nodes, we define ${\rm LB}(\mathcal{J})$:

\begin{equation}
\label{eq-4-2}
    {\rm LB}(\mathcal{J})=\begin{bmatrix}
    \mathcal{J}_{1}^{\top} \\
    \mathcal{J}_{2}^{\top} \\
    \vdots     \\
    \mathcal{J}_{N}^{\top}
    \end{bmatrix}
    =
    \begin{bmatrix}
    \mathcal{J}_{1 1} & \mathcal{J}_{1 2} & \cdots & \mathcal{J}_{1 F^{\text{out}}} \\
    \mathcal{}J_{2 1} & \mathcal{J}_{2 2} & \cdots & \mathcal{J}_{2 F^{\text{out}}} \\
    \vdots     \\
    \mathcal{J}_{N 1} & \mathcal{J}_{N 2} & \cdots & \mathcal{J}_{N F^{\text{out}}}
    \end{bmatrix}_{N \times F^{\text{out}}}
    .
\end{equation}
where $\mathcal{J}_{i j } = \norm{\mJ_{i j}}$.
Based on the definition of ${\rm LB}(\mathcal{J})$, we can further establish the Lipschitz bound of the entire GNN model during training in the next subsection. To measure the scale of ${\rm LB}(\mathcal{J})$ of GNN $f(\cdot)$, we define a $\Lip(f)$ that satisfies
\begin{equation}
\label{eq:jaco:-1}
\Lip(f) = \norm{{\rm LB}(\mathcal{J})}_{\infty, 2}.
\end{equation}
This calculation involves taking the $l_{2}$-$norm$ for each row of $\norm{{\rm LB}(\mathcal{J})}_{\infty, 2}$ and then taking the infinite norm for the entire $\norm{{\rm LB}(\mathcal{J})}_{\infty, 2}$. Now we have proposed an easy solution as Equation~(\ref{eq:jaco:-1}) to approximate $\prod_{l=1}^{L} \left\| {F^{l}}^{\prime} \right\| \left\| \left[\mathcal{J}({h}^{l}) \right]_j\right\|_{\infty}$ for facilitating its use in practical training. 

\subsection{Shed Light on the GNN Fairness: A Rank-Based Perspective}
\label{sec:alg}
\begin{wrapfigure}[17]{R}{0.50\textwidth}
    \begin{minipage}{0.50\textwidth}
    \vspace{-0.2in}
    \begin{algorithm}[H]
    \caption{\footnotesize {JacoLip}: A simplified PyTorch-style Pseudocode of our Lipschitz Bounds for fairness.}\textbf{}
    \label{alg:lip-fair}
  \begin{lstlisting}[language=python]
  # model: graph neural network model
  # Train model for N epochs
  for X, A, target in dataloader_mlp:
     pred = model(X, A)
     ce_loss = CrossEntropyLoss(pred, target)

     # Compute Lipschitz constant for input
     jacobian = Jaco(X, target)
     model_lip = Lip(jacobian) # Eq.(8)
     global_lip = norm(model_lip) # Eq.(9)

     # Optimize model with Lipschitz bound 
     loss = ce_loss + u * global_lip
     loss.backward()
     optimizer.step()
  \end{lstlisting}
    \end{algorithm}
    \vspace{-0.53in}
    \end{minipage}
\end{wrapfigure}
Regularizing GNN training with input-output rank consistency, especially with regard to \textit{rank-based individual fairness}, can be facilitated by employing the Lipschitz bounds of GNNs, denoted as ${\rm Lip}({f})=\max_{j}\prod_{l=1}^{L} \norm{ {F^{l}}^{\prime} } \norm{ \left[\mathcal{J}({h}^{l}) \right]_j}_{\infty}$. The aim is to alleviate biases inherent in training data and promote individual fairness by enforcing Lipschitz constraints on the model output, against the perturbations to final predictions induced by biases learned layer-by-layer from the input across forwarding, which protect the \textit{consistency} between the ranking lists based on the similarity of each node in the input graph to other nodes in the oracle pairwise similarity matrix $\mS_{\mG}$ and the similarity matrix of the predicted outcome space $\mS_{\mY}$, as defined in \textit{Rank-based Individual Fairness of GNNs}, Section~\ref{sec:rank-based-fairness}.

We introduce a plug-and-play solution, named {JacoLip}, that seamlessly integrates with existing fairness-oriented GNN training processes: Algorithm \ref{alg:lip-fair} displays a PyTorch-style pseudocode of this solution in detail. It involves training the GNN model with Lipschitz bounds for a predetermined number of epochs. Throughout the training phase, the Lipschitz constant for the model output is computed using the gradients and norms of the input features. This Lipschitz bound serves as a regularization term within the loss function, ensuring that the model's output stays within defined constraints.
This technique provides a practical way to address individual biases and promote fairness in GNNs without incurring significant computational or memory overheads.

\section{Experiments}
In this section, we conducted two major experiments to examine the Lipschitz bound we analyzed in the previous section:
\textit{(1)} we utilize Lipschitz constants to bound GNN output consistency for increasing rank-based individual fairness on node classification and link prediction tasks;
\textit{(2)} we then validate the constraint effects of Lipschitz bounds on GNN gradients/weights with regards to biases induced by data from a training dynamics perspective.

\subsection{Setup}
\label{sec:exp:setup}
\paragraph{Datasets}
We evaluate the effectiveness of the Lipschitz bound in promoting individual fairness in GNNs from a ranking perspective by conducting experiments on three real-world datasets, each for a chosen downstream task (node classification or link prediction). 
Specifically, we use one citation network (\texttt{ACM}~\citep{tang2008arnetminer}) and two co-authorship networks (\texttt{Co-author-CS} and \texttt{Co-author-Phy}~\citep{shchur2018pitfalls} from the KDD Cup 2016 challenge) for the node classification task. 
For the link prediction task, we use three social networks (\texttt{BlogCatalog}~\citep{tang2009relational}, \texttt{Flickr}~\citep{huang2017label}, and \texttt{Facebook}~\citep{leskovec2012learning}. 
We follow their public train/val/test splits provided by a prior rank-based individual fairness work~\citep{dong2021individual}. 
Detailed statistics of these datasets, as well as the social or real-world meaning of the elements in the graph datasets, are presented in Appendix~\ref{app:dataset}.

\paragraph{Backbones}
We employ two widely-used GNN architectures as backbone models for each downstream learning task in our experiments. Specifically, for the node classification task, we adopt Graph Convolutional Network (GCN)~\citep{kipf2016semi} and Simplifying Graph Convolutional Network (SGC)~\citep{wu2019simplifying}. For the link prediction task, we use GCN and Variational Graph Auto-Encoders (GAE)~\citep{kipf2016variational}. A detailed model card that records configurations of hyperparameters is provided in Appendix~\ref{app:hyper}.

\paragraph{Baselines} In the previous work on rank-based individual fairness~\citep{dong2021individual}, existing group fairness graph embedding methods, such as \citep{bose2019compositional, ijcai2019p456}, are unsuitable for comparison as they promote fairness for subgroups determined by specific protected attributes, whereas our focus is on individual fairness without such attributes. To evaluate our proposed method against this notion of individual fairness, we compare it with three important baselines for rank-based individual fairness:
\begin{itemize}[leftmargin=*]
\item \textit{Redress}~\citep{dong2021individual}: This method proposes a rank-based framework to enhance the individual fairness of GNNs. It integrates GNN model utility maximization and rank-based individual fairness promotion in a joint framework to enable end-to-end training.  
\item \textit{InFoRM}~\citep{kang2020inform}: InFoRM is an individual fairness framework for conventional graph mining tasks, such as PageRank and Spectral Clustering, based on the Lipschitz condition. We adapt InFoRM to different GNN backbone models by combining its individual fairness promotion loss and the unity loss of the GNN backbone model, and optimizing the final loss in an end-to-end manner.
\item \textit{PFR}~\citep{pfr}: PFR aims to learn fair representations to achieve individual fairness. It outperforms traditional approaches, such as \citep{NIPS2016_9d268236, pmlr-v28-zemel13, lahoti2019ifair}, in terms of individual fairness promotion. As PFR can be considered a pre-processing strategy and is not tailored for graph data, we use it on the input node features to generate a new fair node feature representation. 
\end{itemize}
\vspace{-2mm}

\paragraph{Evaluation Metrics} To provide a comprehensive evaluation of rank-based individual fairness, we use two key metrics: the classification accuracy \textit{Acc.} for the node classification task, and the area under the receiver operating characteristic curve \textit{AUC} for the link prediction task. For the individual fairness evaluation, we use a widely used ranking metric following the previous work~\citep{dong2021individual}: \textit{NDCG@k}~\citep{jarvelin2002cumulated}. This metric allows us to measure the similarity between the rankings generated from $\mS_{\mY}$ (result similarity matrix) and $\mS_{\mG}$ (Oracle similarity matrix) for each node. We report the average values of NDCG@k across all nodes and set $k=10$ for quantitative performance comparison. 
\vspace{-2mm}

\paragraph{Implementation Details}
The implementation of all experiments is carried out in PyTorch~\citep{10.5555/3454287.3455008}. We use the released implementations of all GNN backbones used in our experiments, including GCN, SGC, and GAE. The learning rate is set to 0.01 for both the node classification and link prediction tasks. For GCN and SGC-based models, we used two layers with 16 hidden units. For GAE-based models, we use three graph convolutional layers, with 32 and 16 hidden units for the first two layers. We optimize all models using the Adam optimizer~\citep{kingma:adam}. 
More details, including dataset splitting and hyperparameter settings, are provided in the Appendix~\ref{app:hyper}.

\subsection{Effect of Lipschitz Bound to Rank-based Individual Fairness on Graphs}
The experiments conducted on real-world graphs demonstrate the effectiveness of the Lipschitz bound in promoting individual fairness in GNNs from a ranking perspective. The results are summarized in Tables \ref{tab:lip-node-ndcg} and \ref{tab:lip-link-ndcg} for node classification and link prediction tasks, respectively.

Overall, in Table \ref{tab:lip-node-ndcg}, our JacoLip shows promising results in achieving a better trade-off between accuracy and fairness compared to the baselines: 
\textit{(i)} When applied to Vanilla models (GCN or SGC), JacoLip improves the fairness performance (NDCG@10) while maintaining comparable accuracy. This demonstrates that optimizing common GNN backbones with the plug-and-play Lipschitz bounds regularization successfully constrains bias during training and promotes a better trade-off between accuracy and fairness;
\textit{(ii)} Furthermore, when applied to the existing fairness-oriented algorithm Redress, a competitive rank-based individual fairness graph learning method, JacoLip with Lipschitz bounds regularization helps to constrain irrelevant biased factors during training and marginally improves the trade-off between accuracy and fairness across different datasets and backbones.

Similar observations can be made for the link prediction task from Table \ref{tab:lip-link-ndcg}, where the performance of Vanilla (GCN or GAE), InFoRM, PFR, Redress, and JacoLip methods is evaluated using AUC for utility and NDCG@10 for fairness.
In both tasks, JacoLip consistently demonstrates competitive or improved performance compared to the baselines, highlighting the effectiveness of the Lipschitz bound in promoting individual fairness on graphs.
\begin{table}[t]
\vspace{-4.5mm}
\caption{Evaluation on node classification task: comparing under accuracy and NDCG. Higher performance in both metrics indicates a better trade-off. Results are in percentages, and averaged values and standard deviations are computed from five runs. The improvement is within brackets.}
\vspace{-1mm}
\label{tab:lip-node-ndcg}
\begin{center}
\resizebox{1\textwidth}{!}{
\begin{tabular}{lcccccc}
\toprule
 \multirow{2}{*}{Data} & \multirow{2}{*}{Model} & \multirow{2}{*}{Fair Alg.} & \multicolumn{2}{c}{Feature Similarity} & \multicolumn{2}{c}{Structural Similarity} \\
 \cmidrule(r){4-7}
 & & & utility: Acc.$\uparrow$ & fairness: NDCG@10$\uparrow$ & utility: Acc.$\uparrow$ & fairness: NDCG@10$\uparrow$ \\
 \hline
 \multirow{12}{*}{\texttt{ACM}} & \multirow{6}{*}{GCN} & Vanilla~\scriptsize{\citep{kipf2016semi}} &$72.49$$\pm0.6$ &$47.33$$\pm1.0$ &$72.49$$\pm0.6$ &$25.42$$\pm0.6$ \\
 & & InFoRM~\scriptsize{\citep{kang2020inform}} &$68.03$$\pm0.3(-6.15\%)$ &$39.79$$\pm0.3(-15.9\%)$ &$69.13$$\pm0.5(-4.64\%)$ &$12.02$$\pm0.4(-52.7\%)$ \\
 & & PFR~\scriptsize{\citep{pfr}} &$67.88$$\pm1.1(-6.36\%)$ &$31.20$$\pm0.2(-34.1\%)$ &$69.00$$\pm0.7(-4.81\%)$ &$23.85$$\pm1.3(-6.18\%)$ \\
 & & Redress~\scriptsize{\citep{dong2021individual}} &$71.75$$\pm0.4(-1.02\%)$ &$49.13$$\pm0.4(+3.80\%)$ &$72.03$$\pm0.9(-0.63\%)$ &$29.09$$\pm0.4(+14.4\%)$ \\
 & & {\cellcolor{citecolor!10}} \textbf{JacoLip} \footnotesize{(on Vanilla)} &$72.37$$\pm0.3(-0.16\%)$ &$49.80$$\pm0.3(+5.26\%)$ &$71.97$$\pm0.3(-0.71\%)$ &$27.91$$\pm0.7(+9.79\%)$ \\
 & & {\cellcolor{citecolor!10}} \textbf{JacoLip} \footnotesize{(on Redress)} &$71.92$$\pm0.2(-0.78\%)$ &$53.62$$\pm0.6(+13.3\%)$ &$72.05$$\pm0.5(-0.60\%)$ &$31.80$$\pm0.4(+25.1\%)$ \\
 \cmidrule(r){3-7}
 & \multirow{6}{*}{SGC} & Vanilla~\scriptsize{\citep{wu2019simplifying}} &$68.40$$\pm1.0$ &$55.75$$\pm1.1$ &$68.40$$\pm1.0$ &$37.18$$\pm0.6$ \\
 & & InFoRM~\scriptsize{\citep{kang2020inform}} &$68.81$$\pm0.5(+0.60\%)$ &$48.25$$\pm0.5(-13.5\%)$ &$66.71$$\pm0.6(-2.47\%)$ &$28.33$$\pm0.6(-23.8\%)$ \\
 & & PFR~\scriptsize{\citep{pfr}} &$67.97$$\pm0.7(-0.62\%)$ &$34.71$$\pm0.1(-37.7\%)$ &$67.78$$\pm0.1(-0.91\%)$ &$37.15$$\pm0.6(-0.08\%)$ \\
 & & Redress~\scriptsize{\citep{dong2021individual}} &$67.16$$\pm0.2(-1.81\%)$ &$58.64$$\pm0.4(+5.18\%)$ &$67.77$$\pm0.4(-0.92\%)$ &$38.95$$\pm{0.1}(+4.76\%)$ \\
 & & {\cellcolor{citecolor!10}} \textbf{JacoLip} \footnotesize{(on Vanilla)} &$73.84$$\pm0.2(+7.95\%)$ &$62.00$$\pm0.2(+11.21\%)$ &$69.28$$\pm0.3(+1.29\%)$ &$38.36$$\pm0.4(+3.17\%)$ \\
 & & {\cellcolor{citecolor!10}} \textbf{JacoLip} \footnotesize{(on Redress)} &$72.36$$\pm0.4(+5.79\%)$ &$69.22$$\pm0.5(+24.16\%)$ &$72.52$$\pm0.5(+6.02\%)$ &$41.07$$\pm0.3(+10.5\%)$ \\
 \hline
 \multirow{12}{*}{ \texttt{CS} } & \multirow{6}{*}{GCN} & Vanilla~\scriptsize{\citep{kipf2016semi}} &$90.59$$\pm0.3$ &$50.84$$\pm1.2$ &$90.59$$\pm0.3$ &$18.29$$\pm0.8$ \\
 & & InFoRM~\scriptsize{\citep{kang2020inform}} &$88.66$$\pm1.1(-2.13\%)$ &$53.38$$\pm1.6(+5.00\%)$ &$87.55$$\pm0.9(-3.36\%)$ &$19.18$$\pm0.9(+4.87\%)$ \\
 & & PFR~\scriptsize{\citep{pfr}} &$87.51$$\pm0.7(-3.40\%)$ &$37.12$$\pm0.9(-27.0\%)$ &$86.16$$\pm0.2(-4.89\%)$ &$11.98$$\pm1.3(-34.5\%)$ \\
 & & Redress~\scriptsize{\citep{dong2021individual}} &$90.70$$\pm0.2(+0.12\%)$ &$55.01$$\pm1.9(+8.20\%)$ &$89.16$$\pm0.3(-1.58\%)$ &$21.28$$\pm0.3(+16.4\%)$ \\
  & & {\cellcolor{citecolor!10}} \textbf{JacoLip} \footnotesize{(on Vanilla)} &$90.68$$\pm0.3(+0.90\%)$ &$55.35$$\pm0.2(+8.87\%)$ &$89.23$$\pm0.5(-1.50\%)$ &$21.82$$\pm0.2(+19.3\%)$ \\
 & & {\cellcolor{citecolor!10}} \textbf{JacoLip} \footnotesize{(on Redress)} &$90.63$$\pm0.3(+0.40\%)$ &$68.20$$\pm0.4(+34.2\%)$ &$89.21$$\pm0.1(-1.52\%)$ &$31.82$$\pm0.4(+74.1\%)$ \\
 \cmidrule(r){3-7}
 & \multirow{6}{*}{SGC} & Vanilla~\scriptsize{\citep{wu2019simplifying}} &$87.48$$\pm0.8$ &$74.00$$\pm0.1$ &$87.48$$\pm0.8$ &$32.36$$\pm0.3$ \\
 & & InFoRM~\scriptsize{\citep{kang2020inform}} &$88.07$$\pm0.1(+0.67\%)$ &$74.29$$\pm0.1(+0.39\%)$ &$88.65$$\pm0.4(+1.34\%)$ &$32.37$$\pm0.4(+0.03\%)$ \\
 & & PFR~\scriptsize{\citep{pfr}} &$88.31$$\pm0.1(+0.94\%)$ &$48.40$$\pm0.1(-34.6\%)$ &$84.34$$\pm0.3(-3.59\%)$ &$28.87$$\pm0.9(-10.8\%)$ \\
 & & Redress~\scriptsize{\citep{dong2021individual}} &$90.01$$\pm0.2(+2.89\%)$ &$76.60$$\pm0.1(+3.51\%)$ &${8 9.3 5}$$\pm{0.1}(+2.14\%)$ &$34.24$$\pm0.2(+5.81\%)$ \\
  & & {\cellcolor{citecolor!10}} \textbf{JacoLip} \footnotesize{(on Vanilla)} &$90.23$$\pm0.2(+3.14\%)$ &$74.63$$\pm0.2(+0.85\%)$ &$89.53$$\pm0.6(+2.34\%)$ &$32.83$$\pm0.3(+1.45\%)$ \\
 & & {\cellcolor{citecolor!10}} \textbf{JacoLip} \footnotesize{(on Redress)} &$90.12$$\pm0.3(+3.02\%)$ &$77.01$$\pm0.1(+4.07\%)$ &$89.80$$\pm0.2(+2.65\%)$ &$34.89$$\pm0.5(+7.82\%)$ \\
 \hline
 \multirow{12}{*}{ \texttt{Phy} } & \multirow{6}{*}{GCN} & Vanilla~\scriptsize{\citep{kipf2016semi}} &$94.81$$\pm0.2$ &$34.83$$\pm1.1$ &$94.81$$\pm0.2$ &$1.57$$\pm0.1$ \\
 & & InFoRM~\scriptsize{\citep{kang2020inform}} &$89.33$$\pm0.8(-5.78\%)$ &$31.25$$\pm0.0(-10.3\%)$ &$94.46$$\pm0.2(-0.37\%)$ &$1.77$$\pm0.0(+12.7\%)$ \\
 & & PFR~\scriptsize{\citep{pfr}} &$89.74$$\pm0.5(-5.35\%)$ &$24.16$$\pm0.4(-30.6\%)$ &$87.26$$\pm0.2(-7.96\%)$ &$1.20$$\pm0.1(-23.6\%)$ \\
 & & Redress~\scriptsize{\citep{dong2021individual}} &$94.63$$\pm0.7(-0.19\%)$ &$43.64$$\pm0.5(+25.3\%)$ &$93.94$$\pm0.3(-0.92\%)$ &${1.9 3}$$\pm{0.1}(+22.9\%)$ \\
  & & {\cellcolor{citecolor!10}} \textbf{JacoLip} \footnotesize{(on Vanilla)} &$94.60$$\pm0.2(-0.22\%)$ &$37.33$$\pm0.5(+7.18\%)$ &$93.99$$\pm0.4(-0.86\%)$ &$1.87$$\pm0.2(+19.1\%)$ \\
 & & {\cellcolor{citecolor!10}} \textbf{JacoLip} \footnotesize{(on Redress)} &$94.50$$\pm0.2(-0.32\%)$ &$49.37$$\pm0.3(+4.17\%)$ &$93.86$$\pm0.9(-1.00\%)$ &$2.72$$\pm0.1(+73.3\%)$ \\
  \cmidrule(r){3-7}
 & \multirow{6}{*}{SGC} & Vanilla~\scriptsize{\citep{wu2019simplifying}} &$94.45$$\pm0.2$ &$49.63$$\pm0.1$ &$94.45$$\pm0.2$ &$3.61$$\pm0.1$ \\
 & & InFoRM~\scriptsize{\citep{kang2020inform}} &$92.01$$\pm0.1(-2.58\%)$ &$43.87$$\pm0.2(-11.6\%)$ &$94.27$$\pm0.3(-0.19\%)$ &$3.64$$\pm0.0(+0.83\%)$ \\
 & & PFR~\scriptsize{\citep{pfr}} &$89.74$$\pm0.3(-4.99\%)$ &$28.54$$\pm0.1(-42.5\%)$ &$89.73$$\pm0.3(-5.00\%)$ &$2.62$$\pm0.1(-27.4\%)$ \\
 & & Redress~\scriptsize{\citep{dong2021individual}} &$94.30$$\pm0.1(-0.16\%)$ &$53.40$$\pm0.1(+7.60\%)$ &$93.94$$\pm0.2(-0.54\%)$ &$4.03$$\pm0.0(+11.6\%)$ \\
  & & {\cellcolor{citecolor!10}} \textbf{JacoLip} \footnotesize{(on Vanilla)} &$94.20$$\pm0.3(-0.26\%)$ &$50.70$$\pm0.4(+2.16\%)$ &$93.58$$\pm0.2(-0.92\%)$ &$3.80$$\pm0.6(+5.26\%)$ \\
 & & {\cellcolor{citecolor!10}} \textbf{JacoLip} \footnotesize{(on Redress)} &$93.28$$\pm0.1(-1.24\%)$ &$59.20$$\pm0.6(+19.3\%)$ &$93.99$$\pm1.1(-0.49\%)$ &$4.30$$\pm0.5(+19.1\%)$ \\
\toprule
\end{tabular}}
\end{center}
\vspace{-5mm}
\end{table}

\begin{table}[htbp]
\caption{Evaluation on link prediction tasks: comparing under AUC and NDCG. 
}
\vspace{-2mm}
\label{tab:lip-link-ndcg}
\begin{center}
\resizebox{1\textwidth}{!}{
\begin{tabular}{lcccccc}
\toprule
 \multirow{2}{*}{Data} & \multirow{2}{*}{Model} & \multirow{2}{*}{Fair Alg.} & \multicolumn{2}{c}{Feature Similarity} & \multicolumn{2}{c}{Structural Similarity} \\
 \cmidrule(r){4-7}
 & & & utility: AUC$\uparrow$ & fairness: NDCG@10$\uparrow$ & utility: AUC$\uparrow$ & fairness: NDCG@10$\uparrow$ \\
 \hline
 \multirow{12}{*}{\texttt{Blog}} & \multirow{6}{*}{GCN} & Vanilla~\scriptsize{\citep{kipf2016semi}} &$85.87$$\pm0.1$ &$16.73$$\pm0.1$ &$85.87$$\pm0.1$ &$32.47$$\pm0.5$ \\
 & & InFoRM~\scriptsize{\citep{kang2020inform}} &$79.85$$\pm0.6(-7.01\%)$ &$15.57$$\pm0.2(-6.93\%)$ &$84.00$$\pm0.1(-2.18\%)$ &$26.18$$\pm0.3(-19.4\%)$ \\
 & & PFR~\scriptsize{\citep{pfr}} &$84.25$$\pm0.2(-1.89\%)$ &$16.37$$\pm0.0(-2.15\%)$ &$83.88$$\pm0.0(-2.32\%)$ &$29.60$$\pm0.4(-8.84\%)$ \\
 & & Redress~\scriptsize{\citep{dong2021individual}} &$86.49$$\pm0.8(+0.72\%)$ &$17.66$$\pm0.2(+5.56\%)$ &${86.25}$$\pm{0.3}(+0.44\%)$ &${34.62}$$\pm{0.7}(+6.62\%)$ \\
 & & {\cellcolor{citecolor!10}}\textbf{JacoLip} \footnotesize{(on Vanilla)} &$86.51$$\pm0.2(+0.74\%)$ &$17.70$$\pm0.6(+5.79\%)$ &$86.90$$\pm0.5(+1.67\%)$ &$35.00$$\pm0.4(+7.79\%)$ \\
 & & {\cellcolor{citecolor!10}} \textbf{JacoLip} \footnotesize{(on Redress)} &$85.91$$\pm0.2(+0.04\%)$ &$18.02$$\pm0.6(+7.71\%)$ &$86.84$$\pm0.5(+1.13\%)$ &$35.85$$\pm0.4(+10.4\%)$ \\
 \cmidrule(r){3-7}
 & \multirow{6}{*}{GAE} & Vanilla~\scriptsize{\citep{kipf2016variational}} &$85.72$$\pm0.1$ &$17.13$$\pm0.1$ &$85.72$$\pm0.1$ &$41.99$$\pm0.4$ \\
 & & InFoRM~\scriptsize{\citep{kang2020inform}} &$80.01$$\pm0.2(-6.66\%)$ &$16.12$$\pm0.2(-5.90\%)$ &$82.86$$\pm0.0(-3.34\%)$ &$27.29$$\pm0.3(-35.0\%)$ \\
 & & PFR~\scriptsize{\citep{pfr}} &$83.83$$\pm0.1(-2.20\%)$ &$16.64$$\pm0.0(-2.86\%)$ &$83.87$$\pm0.1(-2.16\%)$ &$35.91$$\pm0.4(-14.5\%)$ \\
 & & Redress~\scriptsize{\citep{dong2021individual}} &$84.67$$\pm0.9(-1.22\%)$ &${1 8.1 9}$$\pm{0.1}(+6.19\%)$ &${8 6.3 6}$$\pm{1.5}(+0.75\%)$ &${4 3.5 1}$$\pm{0.7}(+3.62\%)$ \\
 & & {\cellcolor{citecolor!10}} \textbf{JacoLip} \footnotesize{(on Vanilla)} &$85.75$$\pm0.4(+0.03\%)$ &$17.96$$\pm0.5(+4.85\%)$ &$85.86$$\pm0.5(+0.16\%)$ &$42.20$$\pm0.3(+0.50\%)$ \\
 & & {\cellcolor{citecolor!10}} \textbf{JacoLip} \footnotesize{(on Redress)} &$85.70$$\pm0.4(-0.02\%)$ &$18.34$$\pm0.5(+7.06\%)$ &$86.31$$\pm0.5(+0.69\%)$ &$43.60$$\pm0.3(+3.83\%)$ \\
 \hline
 \multirow{12}{*}{\texttt{Flickr}} & \multirow{6}{*}{GCN} & Vanilla~\scriptsize{\citep{kipf2016variational}} &$92.20$$\pm0.3$ &$13.10$$\pm0.2$ &$92.20$$\pm0.3$ &$22.35$$\pm0.3$ \\
 & & InFoRM~\scriptsize{\citep{kang2020inform}} &$91.39$$\pm0.0(-0.88\%)$ &$11.95$$\pm0.1(-8.78\%)$ &$91.73$$\pm0.1(-0.51\%)$ &$23.28$$\pm0.6(+4.16\%)$ \\
 & & PFR~\scriptsize{\citep{pfr}} &$91.91$$\pm0.1(-0.31\%)$ &$12.94$$\pm0.0(-1.22\%)$ &$91.86$$\pm0.2(-0.37\%)$ &$19.80$$\pm0.4(-11.4\%)$ \\
 & & Redress~\scriptsize{\citep{dong2021individual}} &$91.38$$\pm0.1(-0.89\%)$ &${13.58}$$\pm{0.3}(+3.66\%)$ &$92.67$$\pm0.2(+0.51\%)$ &${28.45}$$\pm{0.5}(+27.3\%)$ \\
  & & {\cellcolor{citecolor!10}} \textbf{JacoLip} \footnotesize{(on Vanilla)} &$92.75$$\pm0.3(+0.59\%)$ &$13.74$$\pm0.4(+4.89\%)$ &$92.54$$\pm0.1(+0.37\%)$ &$26.61$$\pm0.4(+19.1\%)$ \\
 & & {\cellcolor{citecolor!10}} \textbf{JacoLip} \footnotesize{(on Redress)} &$92.53$$\pm0.3(+0.35\%)$ &$14.37$$\pm0.4(+9.69\%)$ &$92.69$$\pm0.1(+0.53\%)$ &$28.65$$\pm0.4(+28.2\%)$ \\
\cmidrule(r){3-7}
 & \multirow{6}{*}{GAE} & Vanilla~\scriptsize{\citep{kipf2016variational}} &$89.98$$\pm0.1$ &$12.77$$\pm0.0$ &$89.98$$\pm0.1$ &$23.58$$\pm0.2$ \\
 & & InFoRM~\scriptsize{\citep{kang2020inform}} &$88.76$$\pm0.7(-1.36\%)$ &$12.07$$\pm0.1(-5.48\%)$ &${91.51}$$\pm{0.2}(+1.70\%)$ &$15.78$$\pm0.3(-33.1\%)$ \\
 & & PFR~\scriptsize{\citep{pfr}} &$90.30$$\pm{0.1}(+0.36\%)$ &$12.12$$\pm0.1(-5.09\%)$ &$90.10$$\pm0.1(+1.33\%)$ &$20.46$$\pm0.3(-13.2\%)$ \\
 & & Redress~\scriptsize{\citep{dong2021individual}} &$89.45$$\pm0.5(-0.59\%)$ &$14.24$$\pm0.1(+11.5\%)$ &$89.52$$\pm0.3(-0.51\%)$ &$29.83$$\pm0.2(+26.5\%)$ \\
  & & {\cellcolor{citecolor!10}} \textbf{JacoLip} \footnotesize{(on Vanilla)} &$89.88$$\pm0.3(-0.11\%)$ &$14.37$$\pm0.1(+12.53\%)$ &$89.95$$\pm0.2(-0.03\%)$ &$28.74$$\pm0.5(+21.9\%)$ \\
 & & {\cellcolor{citecolor!10}} \textbf{JacoLip} \footnotesize{(on Redress)} &$89.92$$\pm0.3(-0.06\%)$ &$14.85$$\pm0.1(+16.29\%)$ &$89.56$$\pm0.2(-0.46\%)$ &$30.04$$\pm0.5(+28.7\%)$ \\
 \hline
 \multirow{12}{*}{\texttt{Facebook}} & \multirow{6}{*}{GCN} & Vanilla~\scriptsize{\citep{kipf2016semi}} &$95.60$$\pm1.7$ &$23.07$$\pm0.2$ &$95.60$$\pm1.7$ &$16.55$$\pm1.1$ \\
 & & InFoRM~\scriptsize{\citep{kang2020inform}} &$90.26$$\pm0.1(-5.59\%)$ &$23.23$$\pm0.3(+0.69\%)$ &${9 6.6 6}$$\pm{0.6}(+1.11\%)$ &$15.18$$\pm0.7(-8.28\%)$ \\
 & & PFR~\scriptsize{\citep{pfr}} &$87.11$$\pm1.2(-8.88\%)$ &$21.83$$\pm0.2(-5.37\%)$ &$94.87$$\pm1.9(-0.76\%)$ &$19.53$$\pm0.5(+18.0\%)$ \\
 & & Redress~\scriptsize{\citep{dong2021individual}} &$96.49$$\pm1.6(+0.93\%)$ &$29.60$$\pm0.1(+28.3\%)$ &$92.66$$\pm0.4(-3.08\%)$ &$27.73$$\pm1.1(+67.5\%)$ \\
  & & {\cellcolor{citecolor!10}} \textbf{JacoLip} \footnotesize{(on Vanilla)} &$96.21$$\pm0.2(+0.63\%)$ &$29.47$$\pm0.3(+27.7\%)$ &$95.46$$\pm0.9(-0.14\%)$ &$26.60$$\pm0.1(+60.7\%)$ \\
 & & {\cellcolor{citecolor!10}} \textbf{JacoLip} \footnotesize{(on Redress)} &$96.11$$\pm0.2(+0.53\%)$ &$30.07$$\pm0.3(+30.3\%)$ &$92.76$$\pm0.9(-2.97\%)$ &$28.64$$\pm0.1(+73.1\%)$ \\
\cmidrule(r){3-7}
 & \multirow{6}{*}{GAE} & Vanilla~\scriptsize{\citep{kipf2016variational}} &$98.54$$\pm0.0$ &$26.75$$\pm0.1$ &$98.54$$\pm0.0$ &$27.03$$\pm0.1$ \\
 & & InFoRM~\scriptsize{\citep{kang2020inform}} &$90.50$$\pm0.4(-8.16\%)$ &$22.77$$\pm0.2(-14.9\%)$ &$95.03$$\pm0.1(-3.56\%)$ &$15.38$$\pm0.2(-43.1\%)$ \\
 & & PFR~\scriptsize{\citep{pfr}} &$96.91$$\pm0.1(-1.65\%)$ &$23.52$$\pm0.1(-12.1\%)$ &$98.28$$\pm0.0(-0.26\%)$ &$22.89$$\pm0.3(-15.3\%)$ \\
 & & Redress~\scriptsize{\citep{dong2021individual}} &$95.98$$\pm1.5(-2.60\%)$ &${28.43}$$\pm{0.3}(+6.28\%)$ &$94.07$$\pm1.7(-4.54\%)$ &$33.53$$\pm0.2(+24.0\%)$ \\
  & & {\cellcolor{citecolor!10}} \textbf{JacoLip} \footnotesize{(on Vanilla)} &$97.40$$\pm0.1(-1.16\%)$ &$27.44$$\pm0.6(+2.58\%)$ &$97.02$$\pm1.1(-1.54\%)$ &$30.90$$\pm0.5(+14.3\%)$ \\
 & & {\cellcolor{citecolor!10}} \textbf{JacoLip} \footnotesize{(on Redress)} &$96.10$$\pm0.1(-2.48\%)$ &$28.46$$\pm0.6(+6.39\%)$ &$94.22$$\pm1.1(-4.38\%)$ &$31.62$$\pm0.5(+17.1\%)$ \\
\toprule
\end{tabular}}
\end{center}
\vspace{-5mm}
\end{table}

\subsection{Impact of Lipschitz Bounds on Training Dynamics}
\vspace{-1mm}
\begin{figure}[t]
\centering
\includegraphics[width=1.0\textwidth]{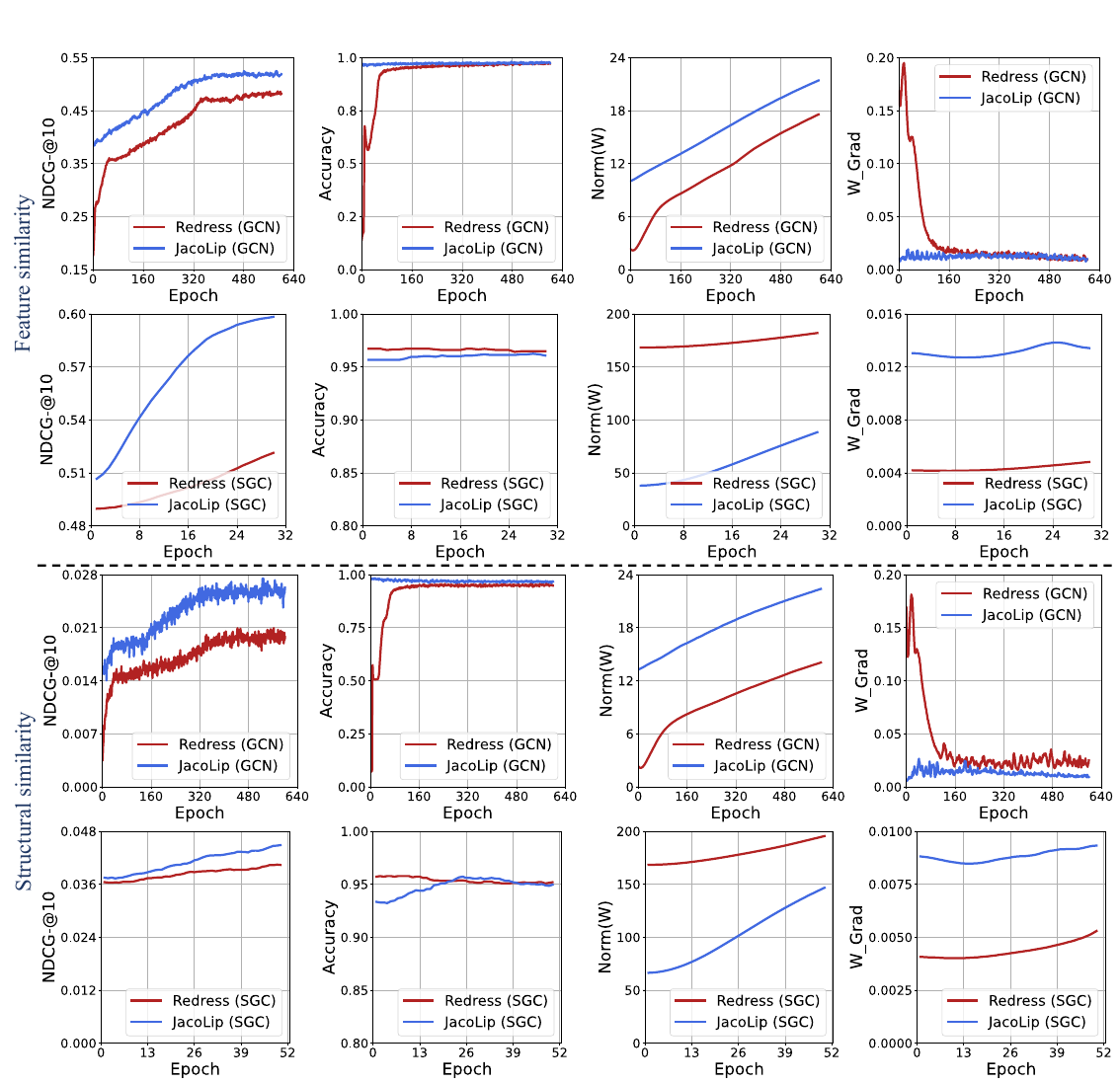}
\vspace{-0.25in}
\caption{
Study of the Lipschitz bounds' impact on model training for rank-based individual fairness. We perform experiments on the co-author-Physics dataset using both nonlinear (GCN) and linear (SGC) models. The training dynamics are assessed by monitoring the NDCG, accuracy, weight norm, and weight gradient as the number of epochs increases.
\textit{Upper two rows}: Metrics under feature similarity;
\textit{Lower two rows}: Metrics under structural similarity.
}
\label{fig:exp}
\vspace{-0.3in}
\end{figure}
We further analyze the impact of Lipschitz bounds through the optimization process and explore its interactions with weight parameters, gradient, fairness, and accuracy during training in Figure~\ref{fig:exp}:

\textit{For the nonlinear GCN model}, our proposed JacoLip demonstrates positive regularizations on gradients. Particularly, at the initial epochs, JacoLip effectively stabilizes gradient magnitudes, leading to higher accuracy (e.g., $\sim$20\% on feature similarity and $\sim$40\% on structural similarity) compared to the baseline Redress. Additionally, during these initial epochs, JacoLip maintains higher fairness metrics NDCG (e.g., $\sim$0.15 on feature similarity) compared to the baseline Redress on GCN;
\textit{For the linear SGC model}, JacoLip also achieves a favorable trade-off between fairness and accuracy compared to the baseline Redress approach. At the start of training, JacoLip on SGC exhibits better fairness (e.g., $\sim$0.2 NDCG on feature similarity) than the baseline Redress on SGC, with only a slight decrease in accuracy. Furthermore, as the number of epochs increases, the accuracy of JacoLip on SGC tends to converge to or even surpass the baseline Redress on SGC.

In summary, the observed accuracy-fairness trade-off during training can be attributed to the constraint effect of Lipschitz bounds on gradient optimization: The higher expressivity of the nonlinear GCN model makes it more prone to losing consistency in the input-output similarity rank, which is crucial for fairness. In contrast, the simplicity of the linear model allows it to preserve consistency more easily, and our JacoLip prioritizes accuracy in this case. These findings offer valuable insights into the dynamic behavior of model training under Lipschitz bounds and highlight the advantages of our proposed JacoLip in promoting fairness while maintaining competitive accuracy.

\vspace{-1mm}
\section{Related Works}
\vspace{-2mm}
\paragraph{Lipschitz Bounds in Deep Models.}
Prior research on Lipschitz constants has primarily focused on specific types of neural networks incorporating convolutional or attention layers \citep{zou2019lipschitz, terris2020building, kim2021lipschitz, araujo2021lipschitz}. In the context of GNNs, \citep{dasoulas2021lipschitz} introduced a Lipschitz normalization method for self-attention layers in GATs. More recently, \citep{gama2022distributed} estimated the filter Lipschitz constant using the infinite norm of a matrix. In contrast, the Lipschitz matrix in our study follows a distinct definition and employs different choices of norm types. Additionally, our objective is to enhance the stability of GNNs against unfair biases, which is not clearly addressed in the aforementioned works.
\vspace{-2mm}
\paragraph{Fair Graph Learning.}
Fair graph learning is a relatively open field~\citep{wu2021learning, dai2021say, buyl2020debayes}. Some existing approaches address fairness concerns through fairness-aware augmentations or adversarial training. 
For instance, Fairwalk \citep{rahman2019fairwalk} is a random walk-based algorithm that aims to mitigate fairness issues in graph node embeddings. Adversarial training is employed in approaches like Compositional Fairness \citep{bose2019compositional} to disentangle learned embeddings from sensitive features. Information Regularization \citep{liao2021information} utilizes adversarial training to minimize the marginal difference between vertex representations. In addition, \citep{48774} improves group fairness by ensuring that node embeddings lie on a hyperplane orthogonal to sensitive features. However, there remains ample room for further exploration in rank-based individual fairness~\citep{dong2021individual}, which is the focus of our work.

\paragraph{Understanding Learning on Graphs.}
Various approaches have emerged to understand the underlying patterns in graph data and its components. Explanatory models for learning on graphs \citep{ying2019gnnexplainer, huang2022graphlime, yuan2021explainability, chen2023characterizing} provide insights into the relationship between a model's predictions and elements in graphs. These works shed light on how local elements or node characteristics influence the decision-making process of GNNs. However, our work differs in that we investigate the impact of Lipschitz constants on practical training dynamics, rather than focusing on trained/fixed-parameter model inference or the influence of local features on GNN decision-making processes.

\section{Conclusions}
\vspace{-1mm}
We have investigated the use of Lipschitz bounds to promote individual fairness in GNNs from a ranking perspective. We conduct a thorough analysis of the theoretical properties of Lipschitz bounds and their relationship to rank-based individual fairness. Building upon this analysis, we propose JacoLip, a Lipschitz-based fairness solution that incorporates Lipschitz bound regularization into the training process of GNNs. To evaluate the effectiveness of JacoLip, extensive experiments are conducted on real-world datasets for node classification and link prediction. The results consistently demonstrate that JacoLip effectively constrains biased factors during training, leading to improved fairness performance while maintaining accuracy.
{
\bibliographystyle{plain}
\bibliography{ref.bib}
}

\newpage
\appendix
\section{Limitations}
While our work provides insights into the Lipschitz bounds of GNNs and their implications for fairness, there are several limitations to be aware of. First, our analysis assumes that the GNN model is well-specified (normally used GCN, SGC, and GAE) and that the data follows certain distributions (citation datasets and co-authorship datasets). Violations of these conditions, such as model misspecification or extremely weird data distributions, may impact the validity of experimental results. Additionally, our discoveries are based on theoretical analysis and empirical evaluations of specific datasets. The generalizability of our results to other domains and datasets remains an open question. Future research should explore these limitations and investigate the impact of these factors on the effectiveness and generalizability of our discoveries.

\section{Broader Impacts}
Our research on Lipschitz bounds for graph neural networks (GNNs) has several potential societal implications, both positive and (potentially) negative: on the positive side, our work contributes to enhancing the stability and interpretability of GNNs, which can lead to more reliable and trustworthy AI systems in various domains. Improved fairness and interpretability of GNNs can help mitigate potential biases and ensure equitable outcomes in decision-making processes; on the (potentially) negative side, the advancement of GNNs also raises concerns about potential misuse and unintended consequences. As with any deep learning technology, GNNs could be deployed in ways that unfairly impact certain groups or perpetuate biases.

To address these challenges, it is crucial to develop better fair training procedures, incorporate strict privacy safeguards, and promote responsible deployment and monitoring of GNNs. Therefore, we contribute to the foundation of GNN research in this work, and it is important for future studies and applications to consider the broader impacts and potential harms while adopting appropriate mitigation strategies.
\section{Proofs}
\label{app:proof}
\subsection{Notations}
We use the following notations throughout the paper. 
Sets are denoted by $\{\}$ and vectors by $( )$. For $n \in \sN$, we denote $[n] = \{1, \cdots, n\}$.
Scalars are denoted by regular letters, lowercase bold letters denote vectors, and uppercase bold letters denote matrices. For instance, $\vx = ( x_1, \cdots, x_n )^\T \in \R^n $ and $\mX = [X_{ik}]_{i \in [n], k \in [m]} \in \R^{n \times m}$.
For any vector $\vx \in \R^n$, we use $\norm{\vx}$ to denote its $\ell_2$-norm: $\norm{\vx} = \left( \sum_{i=1}^n x_i^2 \right)^{1/2}$. For any matrix $\mX \in \R^{n \times m}$, we use $\mX_{i,:}$ to denote its $i$-th row and $\mX_{:,k}$ to denote its $k$-th column. The $(\infty, 2)$-norm of $\mX$ is denoted by $\norm{\mX}_{\infty, 2} = \max_{i \in [n]} \norm{\mX_{i,:}}$.
Given a graph $G = G(V,E)$ with ordered nodes, we denote its adjacency matrix by $\mA$ such that $A_{ij} = 1$ if $\{i,j\} \in E$ and $A_{ij} = 0$ otherwise. When it is clear from the context, we use $\mX \in \R^{N \times F}$ to denote a feature matrix whose $i$-th row corresponds to the features of the $i$-th node, and the $j$-th column represents the features across all nodes for the $j$-th attribute. We denote the output of the GNN as $\mY \in \R^{N \times C}$, where $N$ is the number of nodes and $C$ is the number of output classes.

\subsection{Proof of Lemma \ref{lemma:inequality} in Section~\ref{sec:stability}}
\label{app:lemma}
\setcounter{lemma}{0} 
\begin{tcolorbox}[width=1.0\linewidth, colback=citecolor!2, colframe=black, arc=0pt, boxsep=0mm, arc=2mm, left=2mm, right=2mm, top=5mm, bottom=2mm]
\vspace{-0.3cm}
\begin{lemma}
\label{lemma:inequality-app}
For any vectors \(x\) and \(y\) in a Euclidean space, let \(g(x)\) and \(g(y)\) be vector-valued functions of \(x\) and \(y\) respectively, and let \(g_i\) be the \(i\)-th component function of \(g\). Then, we have the following inequality:
\begin{equation}
\label{lemma-eq}
    \frac{\left\|g(x)-g(y)\right \|}{\|x-y\|} \leqslant \left\| \left[\frac{g_i(x)-g_i(y)}{\|x-y\|}\right]_{i=1}^n \right\|,
\end{equation}
\end{lemma}
\end{tcolorbox}
Lemma \ref{lemma:inequality} provides an inequality that relates the norm of the difference between two vector-valued functions, \(g(x)\) and \(g(y)\), to the norm of a vector composed of the component-wise differences of the functions evaluated at \(x\) and \(y\). Based on Lemma \ref{lemma:inequality}, we can now present the following theorem:
\begin{proof}
We begin by observing that
    \begin{equation}
    \begin{aligned}
        \frac{\norm{g(\vx)-g(\vy)}}{\norm{\vx-\vy}} = & \frac{\norm{\left[ g_{i}(\vx)-g_{i}(\vy) \right]_{i=1}^n}}{\norm{\vx-\vy}} \\
        = & \norm{\left[ \frac{ \abs{ g_{i}(\vx)-g_{i}(\vy) } }{\norm{\vx-\vy}}\right]_{i=1}^n}.
    \end{aligned}
    \end{equation}
Furthermore, for each $i \in [n]$, $\displaystyle \frac{\abs{g_{i}(\vx)-g_{i}(\vy)}}{\norm{\vx - \vy}}\le {\rm Lip}(g_{i})$. Therefore, we can write
    \begin{equation}
    \begin{split}
        {\rm Lip}(g) &= \sup_{\vx \neq \vy} \frac{\norm{g(\vx)-g(\vy)}}{\norm{\vx-\vy}} \\
        & = \sup_{\vx \neq \vy} \norm{\left[ \frac{ \abs{ g_{i}(\vx)-g_{i}(\vy) } }{\norm{\vx-\vy}}\right]_{i=1}^n} \\
        & \le \sup_{\vx \neq \vy} \norm{ \left[ {\rm Lip}(g_i) \right]_{i=1}^n } = \norm{ \left[ {\rm Lip}(g_i) \right]_{i=1}^n },
    \end{split}
    \end{equation}
this completes the proof.
\end{proof}
In the above proof of Lemma \ref{lemma:inequality}, we start by rewriting the norm of the difference between \(g(\vx)\) and \(g(\vy)\) divided by the norm of \(\vx-\vy\) as a norm of a vector containing the component-wise differences of \(g_i(\vx)\) and \(g_i(\vy)\) divided by the norm of \(\vx-\vy\) for each \(i\). We then observe that for each \(i\), the absolute value of \(g_i(\vx)-g_i(\vy)\) divided by the norm of \(\vx-\vy\) is bounded by the Lipschitz constant \({\rm Lip}(g_i)\). Hence, the Lipschitz constant of \(g\) is bounded by the norm of the vector \(\left[ {\rm Lip}(g_i) \right]_{i=1}^n\). This establishes the inequality in Lemma \ref{lemma:inequality}.

\subsection{Proof of Theorem \ref{thm-1} in Section~\ref{sec:stability}}
\label{app:proof-app}
\setcounter{theorem}{0} 

\begin{tcolorbox}[width=1.0\linewidth, colback=citecolor!2.5, colframe=black, arc=0pt, boxsep=0mm, arc=2mm, left=2mm, right=2mm, top=5mm, bottom=2mm]
\vspace{-0.3cm}
\begin{theorem}
\label{thm-1-1}
Let $\mY$ be the output of an $L$-layer GNN (represented in $f(\cdot)$) with $\mX$ as the input. Assuming the activation function (represented in $\rho(\cdot)$) is ReLU with a Lipschitz constant of ${\rm Lip}({\rho}) = 1$, then the global Lipschitz constant of the GNN, denoted as ${\rm Lip}({f})$, satisfies the following inequality:
\begin{equation}
\label{eq:thm-1-app}
{\rm Lip}({f}) \leqslant \max_j \prod_{l=1}^{L} \left\| {F^{l}}^{\prime} \right\| \left\| \left[\mathcal{J}({h}^{l}) \right]_{j}\right\|_{\infty},
\end{equation}
where ${F^{l}}^{\prime}$ represents the output dimension of the $l$-th message-passing layer, $j$ is the index of the node (e.g., $j$-th), and the vector $\left[\mathcal{J}({h}^{l}) \right] = \left[\norm{\mJ_{1}({h}^{l})}, \norm{\mJ_{2}({h}^{l})}, \cdots, \norm{\mJ_{{F^{l}}^{\prime}}({h}^{l})}\right]$. Notably, $\mJ_i(h^{l})$ denotes the $i$-th row of the Jacobian matrix of the $l$-th layer's input and output, and $\left[\mathcal{J}({h}^{l}) \right]_{j}$ is the vector corresponding to the $j$-th node in the $l$-th layer ${h}^{l}(\cdot)$.
\end{theorem}
\end{tcolorbox}

Theorem \ref{thm-1} provides an inequality that bounds the global Lipschitz constant of the GNN based on the layer outputs and Jacobian matrices. It is derived as follows:
\begin{proof}
We begin by examining the Lipschitz property of the GNN. Let $\mY$ denote the output of an $L$-layer GNN with input $\mX$. Assuming the commonly used ReLU activation function as the non-linear layer ${\rho}(\cdot)$, we have ${\rm Lip}({\rho}) = 1$.
First, we consider the Lipschitz constant between the hidden states of two nodes output by any message-passing layer ${h}(\cdot)$ in ${f}(\cdot)$. Let $z_1$ and $z_{2}$ represent the hidden states of node features $x_1$ and $x_2$, respectively. The Lipschitz constant between these hidden states is given by:
\begin{equation}
\frac{\left\|z_{1} - z_{2}\right\|}{\left\|x_1 - x_2\right\|} = \frac{\|\left({h}(x_1) - {h}(x_2)\right)\|}{\left\|x_1 - x_2\right\|}.
\end{equation}
By applying the triangle inequality, we obtain:
\begin{equation}
\frac{\left\|z_{1} - z_{2}\right\|}{\left\|x_1 - x_2\right\|} \leqslant \left\|\left[\frac{{h}(x_1)_{i} - {h}(x_2)_{i}}{\left\|x_1 - x_2\right\|}\right]_{i=1}^{F^{\prime}}\right\|,
\end{equation}
Next, we consider the Lipschitz constant between individual elements of the hidden states. Let ${f}(x_1)$ and ${f}(x_2)$ represent the hidden state matrices for inputs $x_1$ and $x_2$, respectively. By again applying the triangle inequality, we have:
\begin{equation}
\frac{\left\|z_{1} - z_{2}\right\|}{\left\|x_1 - x_2\right\|} \leqslant \left\| F^{\prime} \times \max_i \frac{{h}(x_1)_i - {h}(x_2)_i}{\left\|x_1 - x_2\right\|} \right\|,
\end{equation}
let's focus on the Lipschitz constant of the individual elements, $\frac{{h}(x_1)_i - {h}(x_2)_i}{\left\|x_1 - x_2\right\|}$: Here, ${f}(x)_i$ denotes the $i$-th column of the matrix ${f}(x)$.
We denote ${h}^{l}(\cdot)$ as the operation of the $l$-th message-passing layer in ${f}(\cdot)$, then by applying the triangle inequality and leveraging the Lipschitz property of the ReLU activation function, we have:
\begin{equation}
\frac{\left\|{f}(x_{1})_{j, 1} - {f}(x_{2})_{j, 2}\right\|}{\left\|x_{j, 1} - x_{j, 2}\right\|} \leqslant \prod_{l=1}^{L} \left\| {F^{l}}^{\prime} \right\| \left\| \left[\mathcal{J}({h}^{l}) \right]_j\right\|_{\infty},
\end{equation}
where $x_{j, 1}$ and $x_{j, 1}$ denote features of $j$-th node's in $x_{1}$ and $x_{2}$, respectively, and $\left[\mathcal{J}({h}^{l}) \right]_j$ represents the $j$-th node's the Jacobian matrix of the $l$-th message-passing layer.
Therefore, the Lipschitz constant for the GNN can be expressed as:
\begin{equation}
{\rm Lip}({f}) = \max_{j}\prod_{l=1}^{L} \left\| {F^{l}}^{\prime} \right\| \left\| \left[\mathcal{J}({h}^{l}) \right]_j\right\|_{\infty}.
\label{eq:lip-bounds}
\end{equation}
In summary, we have shown that for any two input samples $x_1$ and $x_2$, the Lipschitz constant of the GNN, denoted as ${\rm Lip}({f})$, satisfies:
\begin{equation}
\label{eq:lip:-2}
\left\| \mY_{1} - \mY_{2} \right\| \leqslant {\rm Lip}({f}) \left\| \mX_{1} - \mX_{2} \right\|,
\end{equation}
where $\mY$ denotes the output of the GNN for inputs $\mX$. This inequality implies that the Lipschitz constant ${\rm Lip}({f})$ controls the magnitude of changes in the output based on input biases/perturbations.
Therefore, we have established the following result:
\begin{equation}
\label{eq:lip:-1}
\left\| \mY_{1} - \mY_{2} \right\| \leqslant \prod_{l=1}^{L} \left\| {F^{l}}^{\prime} \right\| \left\| \left[\mathcal{J}({h}^{l}) \right]_j\right\|_{\infty} \left\| \mX_{1} - \mX_{2} \right\|.
\end{equation}
This inequality demonstrates that the Lipschitz constant of the GNN, ${\rm Lip}({f})$, controls the magnitude of the difference in the output $\mY$ based on the difference in the input $\mX$. It allows us to analyze the stability of the model's output with respect to input perturbations.
\end{proof}

\section{Additional Experiments}
\label{app:sec:exp}

According to Table~\ref{tab:lip-node-err}, on node classification tasks, JacoLip consistently shows a competitive or improved trade-off between accuracy and error compared to the baselines, highlighting the effectiveness of the Lipschitz bound in promoting individual fairness on graphs.

\begin{table}[t]
\caption{Evaluation on node classification tasks: comparing under accuracy and error.}
\label{tab:lip-node-err}
\begin{center}
\resizebox{1\textwidth}{!}{
\begin{tabular}{lcccccc}
\toprule
 \multirow{2}{*}{Data} & \multirow{2}{*}{Model} & \multirow{2}{*}{Fair Alg.} & \multicolumn{2}{c}{Feature Similarity} & \multicolumn{2}{c}{Structural Similarity} \\
 \cmidrule(r){4-7}
 & & & utility: Acc.$\uparrow$ & fairness: Err.@10$\uparrow$ & utility: Acc.$\uparrow$ & fairness: Err.@10$\uparrow$ \\
 \hline
 \multirow{12}{*}{ \texttt{ACM} } & \multirow{6}{*}{GCN} & Vanilla~\scriptsize{\citep{kipf2016semi}} &$72.49$$\pm0.6$ &$75.70$$\pm0.6$ &$72.49$$\pm0.6$ &$37.55$$\pm0.4$ \\
 & & InFoRM~\scriptsize{\citep{kang2020inform}} &$67.65$$\pm1.0{(-6.68\%)}$ &$73.49$$\pm0.5(-2.92\%)$ &$65.91$$\pm0.2(-9.07\%)$ &$19.96$$\pm0.6(-46.8\%)$ \\
 & & PFR~\scriptsize{\citep{pfr}} &$68.48$$\pm0.6(-5.53\%)$ &$76.28$$\pm0.1(0.77\%)$ &$70.22$$\pm0.7(-3.13\%)$ &$36.54$$\pm0.4(-2.69\%)$ \\
 & & Redress~\scriptsize{\citep{dong2021individual}} &$73.46$$\pm0.2(+1.34\%)$ &$82.27$$\pm0.1(+8.68\%)$ &$71.87$$\pm0.4(-0.86\%)$ &$43.74$$\pm0.0(+16.5\%)$ \\
 & & {\cellcolor{citecolor!10}} \textbf{JacoLip} \footnotesize{(on Vanilla)} &$72.80$$\pm0.2(+4.27\%)$ &$82.88$$\pm0.1(+9.48\%)$ &$72.30$$\pm0.4(-0.26\%)$ &$39.28$$\pm0.2(+4.61\%)$ \\
 & & {\cellcolor{citecolor!10}} \textbf{JacoLip} \footnotesize{(on Redress)} &$71.05$$\pm0.4(+2.00\%)$ &$82.21$$\pm0.3(+8.60\%)$ &$71.92$$\pm0.3(-0.79\%)$ &$46.13$$\pm0.3(+22.85\%)$ \\
 \cmidrule(r){3-7}
 & \multirow{6}{*}{SGC} & Vanilla~\scriptsize{\citep{wu2019simplifying}} &$68.40$$\pm1.0$ &$80.06$$\pm0.1$ &$68.40$$\pm1.0$ &$45.95$$\pm0.3$ \\
 & & InFoRM~\scriptsize{\citep{kang2020inform}} &$67.96$$\pm0.5(-0.64\%)$ &$75.63$$\pm0.5(-5.53\%)$ &$66.16$$\pm0.6(-3.27\%)$ &$39.79$$\pm0.1(-13.4\%)$ \\
 & & PFR~\scriptsize{\citep{pfr}} &$67.69$$\pm0.4(-1.04\%)$ &$76.80$$\pm0.1(-4.07\%)$ &$66.69$$\pm0.3(-2.50\%)$ &$46.99$$\pm0.5(+2.26\%)$ \\
 & & Redress~\scriptsize{\citep{dong2021individual}} &$66.51$$\pm0.3(-2.76\%)$ &$82.32$$\pm0.3(+2.82\%)$ &$67.10$$\pm0.7(-1.90\%)$ &$49.02$$\pm0.2(+4.76\%)$ \\
 & & {\cellcolor{citecolor!10}} \textbf{JacoLip} \footnotesize{(on Vanilla)} &$74.04$$\pm0.2(+8.25\%)$ &$82.73$$\pm0.7(+5.18\%)$ &$72.91$$\pm0.9(+3.33\%)$ &$48.64$$\pm0.2(+5.85\%)$ \\
 & & {\cellcolor{citecolor!10}} \textbf{JacoLip} \footnotesize{(on Redress)} &$69.91$$\pm0.1(+2.21\%)$ &$85.22$$\pm0.4(+6.45\%)$ &$71.27$$\pm0.3(+4.20\%)$ &$52.0$$\pm0.4(+13.23\%)$ \\
 \hline
 \multirow{12}{*}{ \texttt{CS} } & \multirow{6}{*}{GCN} & Vanilla~\scriptsize{\citep{kipf2016semi}} &$90.59$$\pm0.3$ &$80.41$$\pm0.1$ &$90.59$$\pm0.3$ &$26.69$$\pm1.3$ \\
 & & InFoRM~\scriptsize{\citep{kang2020inform}} &$88.37$$\pm0.9(-2.45\%)$ &$80.63$$\pm0.6(+0.27\%)$ &$87.10$$\pm0.9(-3.85\%)$ &$29.68$$\pm0.6(+11.2\%)$ \\
 & & PFR~\scriptsize{\citep{pfr}} &$87.62$$\pm0.2(-3.28\%)$ &$76.26$$\pm0.1(-5.16\%)$ &$85.66$$\pm0.7(-5.44\%)$ &$19.80$$\pm1.4(-25.8\%)$ \\
 & & Redress~\scriptsize{\citep{dong2021individual}} &$90.06$$\pm0.5(-0.59\%)$ &$83.24$$\pm0.2(+3.52\%)$ &$89.91$$\pm0.2(-0.86\%)$ &$32.42$$\pm1.6(+21.5\%)$ \\
  & & {\cellcolor{citecolor!10}} \textbf{JacoLip} \footnotesize{(on Vanilla)} &$90.41$$\pm0.4(-0.20\%)$ &$82.57$$\pm0.1(+2.69\%)$ &$89.12$$\pm0.1(-1.62\%)$ &$32.8$$\pm0.6(+22.74\%)$ \\
 & & {\cellcolor{citecolor!10}} \textbf{JacoLip} \footnotesize{(on Redress)} &$90.30$$\pm0.3(-0.32\%)$ &$88.11$$\pm0.3(+9.58\%)$ &$89.93$$\pm0.2(-0.73\%)$ &$42.5$$\pm0.4(+59.24\%)$ \\
 \cmidrule(r){3-7}
 & \multirow{6}{*}{SGC} & Vanilla~\scriptsize{\citep{wu2019simplifying}} &$87.48$$\pm0.8$ &$90.58$$\pm0.1$ &$87.48$$\pm0.8$ &$43.28$$\pm0.2$ \\
 & & InFoRM~\scriptsize{\citep{kang2020inform}} &$87.31$$\pm0.5(-0.19\%)$ &$90.64$$\pm0.1(+0.07\%)$ &$88.21$$\pm0.4(+0.83\%)$ &$44.37$$\pm0.1(+0.21\%)$ \\
 & & PFR~\scriptsize{\citep{pfr}} &$87.95$$\pm0.2(+0.54\%)$ &$79.85$$\pm0.2(-11.8\%)$ &$86.93$$\pm0.1(-0.63\%)$ &$38.83$$\pm0.8(-10.3\%)$ \\
 & & Redress~\scriptsize{\citep{dong2021individual}} &$90.48$$\pm0.2(+3.43\%)$ &$92.03$$\pm0.1(+1.60\%)$ &$ 90.39$$\pm0.1(+3.33\%)$ &$45.81$$\pm0.0(+5.85\%)$ \\
  & & {\cellcolor{citecolor!10}} \textbf{JacoLip} \footnotesize{(on Vanilla)} &$90.71$$\pm0.3(+3.69\%)$ &$90.75$$\pm0.4(+0.19\%)$ &$90.34$$\pm1.0(+3.27\%)$ &$43.92$$\pm0.3(+1.48\%)$ \\
 & & {\cellcolor{citecolor!10}} \textbf{JacoLip} \footnotesize{(on Redress)} &$92.22$$\pm0.2(+5.42\%)$ &$92.22$$\pm0.4(+1.81\%)$ &$90.54$$\pm0.3(+3.50\%)$ &$46.39$$\pm0.5(+7.19\%)$ \\
 \hline
 \multirow{12}{*}{ \texttt{Phy} } & \multirow{6}{*}{GCN} & Vanilla~\scriptsize{\citep{kipf2016semi}} &$94.81$$\pm0.2$ &$73.25$$\pm0.3$ &$94.81$$\pm0.2$ &$2.58$$\pm0.1$ \\
 & & InFoRM~\scriptsize{\citep{kang2020inform}} &$88.67$$\pm0.7(-6.48\%)$ &$73.80$$\pm0.6(+0.75\%)$ &$94.68$$\pm0.2(-0.14\%)$ &$2.45$$\pm0.1(-5.04\%)$ \\
 & & PFR~\scriptsize{\citep{pfr}} &$88.79$$\pm0.2(-6.35\%)$ &$73.22$$\pm0.4(+0.10\%)$ &$89.69$$\pm1.0(-5.40\%)$ &$1.67$$\pm0.1(-35.3\%)$ \\
 & & Redress~\scriptsize{\citep{dong2021individual}} &$93.71$$\pm0.1(-1.16\%)$ &$80.23$$\pm0.1(+9.53\%)$ &$93.91$$\pm0.4(-0.95\%)$ &$ 3.22$$\pm0.3(+22.9\%)$ \\
  & & {\cellcolor{citecolor!10}} \textbf{JacoLip} \footnotesize{(on Vanilla)} &$93.71$$\pm0.2(-1.16\%)$ &$78.64$$\pm1.1(+7.36\%)$ &$94.75$$\pm0.3(-0.06\%)$ &$2.75$$\pm0.6(+6.80\%)$ \\
 & & {\cellcolor{citecolor!10}} \textbf{JacoLip} \footnotesize{(on Redress)} &$93.79$$\pm0.8(-1.08\%)$ &$82.6$$\pm0.3(+12.70\%)$ &$93.98$$\pm0.3(-0.88\%)$ &$4.0$$\pm0.1(+55.43\%)$ \\
  \cmidrule(r){3-7}
 & \multirow{6}{*}{SGC} & Vanilla~\scriptsize{\citep{wu2019simplifying}} &$94.45$$\pm0.2$ &$77.48$$\pm0.2$ &$94.45$$\pm0.2$ &$4.50$$\pm0.1$ \\
 & & InFoRM~\scriptsize{\citep{kang2020inform}} &$92.06$$\pm0.2(-2.53\%)$ &$75.13$$\pm0.4(-3.03\%)$ &$94.27$$\pm0.1(-0.19\%)$ &$4.44$$\pm0.0(-1.33\%)$ \\
 & & PFR~\scriptsize{\citep{pfr}} &$87.39$$\pm1.2(-7.47\%)$ &$73.42$$\pm0.2(-5.24\%)$ &$89.16$$\pm0.3(-5.60\%)$ &$3.41$$\pm0.2(-24.2\%)$ \\
 & & Redress~\scriptsize{\citep{dong2021individual}} &$94.81$$\pm0.2(+0.38\%)$ &$79.57$$\pm0.2(+2.70\%)$ &$94.54$$\pm0.1(+0.10\%)$ &$4.98$$\pm0.1(+10.7\%)$ \\
  & & {\cellcolor{citecolor!10}} \textbf{JacoLip} \footnotesize{(on Vanilla)} &$94.43$$\pm0.7(-0.02\%)$ &$78.82$$\pm0.8(+1.73\%)$ &$94.09$$\pm0.6(-0.38\%)$ &$4.75$$\pm0.2(+5.56\%)$ \\
 & & {\cellcolor{citecolor!10}} \textbf{JacoLip} \footnotesize{(on Redress)} &$94.78$$\pm0.1(+0.35\%)$ &$82.21$$\pm0.2(+6.10\%)$ &$93.00$$\pm1.3(-1.54\%)$ &$5.45$$\pm0.1(+1.90\%)$ \\
\toprule
\end{tabular}}
\end{center}
\end{table}

\section{Model Card}
\label{app:hyper}
\subsection{Implementations}
Code and datasets are available anonymously at \url{https://tinyurl.com/JacoLip}.

\subsection{Hyperparameters Configurations}
The hyper-parameters for our method across all datasets are listed in Table~\ref{tab:hyper}. For fair comparisons, we follow the default settings of Redress~\citep{dong2021individual}.
\begin{table}[H]
    \centering
    \caption{Hyperparameters used in our experiments.}
    \resizebox{0.95\columnwidth}{!}{
    \begin{tabular}{lcccccc}
    \toprule 
    \multirow{2}{*}{Hyperparameters} & \multicolumn{3}{c}{Node Classification} & \multicolumn{3}{c}{Link Prediction} \\
        \cmidrule(r){2-7} 
    & \texttt{ACM} & \texttt{Coauthor-CS} & \texttt{Coauthor-Phy} & \texttt{Blog} & \texttt{Flickr} & \texttt{Facebook}\\
    \cmidrule(r){1-7} 
     \multicolumn{7}{c}{Hyperparameters w.r.t. the GCN model}\\
     \cmidrule(r){1-7} 
     \# Layers         & $2$ & $2$ & $2$ & $2$ & $2$ & $2$ \\
     Hidden Dimension  & $[16, 9]$ & $[16, 15]$ & $[16, 5]$ & $[32, 16]$ & $[32, 16]$ & $[32, 16]$ \\
     Activation   & \multicolumn{6}{c}{ReLU used for all datasets} \\
     Dropout  & $0.03$ & $0.03$ & $0.03$ & $0.00$ & $0.00$ & $0.00$  \\
     Optimizer    & \multicolumn{6}{c}{AdamW with $1e-5$ weight decay used for all datasets } \\
     Pretrain Steps& $300$ & $300$ & $300$ & $200$ & $200$ & $200$ \\
     Training Steps& $150$ & $200$ & $200$ & $60$ &  $100$ & $50$ \\
     Learning Rate & $0.01$ & $0.01$ & $0.01$ & $0.01$ & $0.01$  & $0.01$  \\
     \cmidrule(r){1-7} 
     \multicolumn{7}{c}{Hyperparameters w.r.t. the SGC model}\\
     \cmidrule(r){1-7} 
     \# Layers         & $1$ & $1$ & $1$ & N.A. & N.A. & N.A. \\
     Hidden Dimension  & N.A. & N.A. & N.A. & N.A. & N.A. & N.A. \\
     Dropout  & N.A. & N.A. & N.A. & N.A. & N.A.  & N.A.  \\
     Optimizer    & \multicolumn{3}{c}{AdamW with $1e-5$ weight decay}  & N.A. & N.A. & N.A. \\
     Pretrain Steps& $300$ & $500$ & $500$ & N.A. & N.A. & N.A. \\
     Training Steps& $15$  & $40$  & $30$  & N.A. & N.A. & N.A. \\
     Learning Rate& $0.01$ & $0.01$ & $0.01$ & N.A. & N.A. & N.A.    \\
     \cmidrule(r){1-7} 
     \multicolumn{7}{c}{Hyperparameters w.r.t. the GAE model}\\
     \cmidrule(r){1-7} 
     \# Layers         & N.A. & N.A. & N.A. & $2$ & $2$ & $2$ \\
     Hidden Dimension  & N.A. & N.A. & N.A. & $[32, 16]$ & $[32, 16]$ & $[32, 16]$ \\
     Activation        & N.A. & N.A. & N.A. & N.A. & N.A. & N.A.  \\
     Dropout           & $0.0$  & $0.0$  & $0.0$  & $0.0$  & $0.0$  & $0.0$  \\
     Optimizer         & N.A. & N.A. & N.A. & \multicolumn{3}{c}{AdamW with $1e-5$ weight decay}   \\
     Pretrain Steps    & N.A. & N.A. & N.A. & $200$ & $200$ & $200$ \\
     Training Steps    & N.A. & N.A. & N.A. & $60$  & $100$ & $50$ \\
     Learning Rate & N.A. & N.A. & N.A. & $0.01$ & $0.01$ & $0.01$ \\
    \bottomrule 
    \end{tabular}}
    \label{tab:hyper}
\end{table}

\section{Dataset Descriptions}
\label{app:dataset}
In this section, we provide additional details about the datasets utilized in our work, as discussed in Section~\ref{sec:exp:setup}:

\paragraph{Citation Networks}
For citation networks, each node corresponds to a paper, and an edge between two nodes represents the citation relationship between the respective papers. The node attributes in these networks are generated using the bag-of-words model based on the abstract sections of the published papers.

\paragraph{Co-author Networks}
Co-author networks consist of nodes representing authors, where an edge between two nodes indicates that the corresponding authors have collaborated on a paper. The node attributes in these networks are constructed based on the bag-of-words model using the authors' profiles or descriptions.

\paragraph{Social Networks}
In social networks, each node represents a user, and the links between nodes represent interactions between users. The attributes associated with these nodes are derived from user profiles or descriptions.

The datasets used in our work are referred to as CS and Phy, which are abbreviations for the Co-author-CS and Co-author-Phy datasets, respectively. A comprehensive overview of the datasets, including their detailed statistics, is presented in Table~\ref{tab:dataset-stats}.

\begin{table}[H]
\centering
\caption{Detailed statistics of the datasets used for node classification and link prediction. We follow the default settings of Redress~\citep{dong2021individual} fair comparisons.}
\label{tab:dataset-stats}
\vspace{2mm}
\resizebox{0.7\textwidth}{!}{
\begin{tabular}{llcccc}
\hline
Task & Dataset & \# Nodes & \# Edges & \# Features & \# Classes \\
\hline
\multirow{3}{*}{node classification} & \texttt{ACM} & $16,484$ & $71,980$ & $8,337$ & $9$ \\
& \texttt{Coauthor-CS} & $18,333$ & $81,894$ & $6,805$ & $15$ \\
& \texttt{Coauthor-Phy} & $34,493$ & $247,962$ & $8,415$ & $5$ \\
\hline
\multirow{3}{*}{link prediction} & \texttt{BlogCatalog} & $5,196$ & $171,743$ & $8,189$ & N.A. \\
& \texttt{Flickr} & $7,575$ & $239,738$ & $12,047$ & N.A. \\
& \texttt{Facebook} & $4,039$ & $88,234$ & $1,406$ & N.A. \\
\hline
\end{tabular}}
\end{table}

\end{document}